\documentclass[sigconf]{acmart}

\usepackage{tabulary}
\usepackage{hyperref}

\usepackage{algorithm}
\usepackage{listings}
\usepackage{algpseudocode}
\usepackage{multirow}
\usepackage{etoolbox}
\makeatletter
\AfterEndEnvironment{algorithm}{\let\@algcomment\relax}
\AtEndEnvironment{algorithm}{\kern2pt\hrule\relax\vskip3pt\@algcomment}
\let\@algcomment\relax
\newcommand\algcomment[1]{\def\@algcomment{\footnotesize#1}}
\renewcommand\fs@ruled{\def\@fs@cfont{\bfseries}\let\@fs@capt\floatc@ruled
  \def\@fs@pre{\hrule height.8pt depth0pt \kern2pt}%
  \def\@fs@post{}%
  \def\@fs@mid{\kern2pt\hrule\kern2pt}%
  \let\@fs@iftopcapt\iftrue}
\makeatother

\AtBeginDocument{%
  \providecommand\BibTeX{{%
    \normalfont B\kern-0.5em{\scshape i\kern-0.25em b}\kern-0.8em\TeX}}}

\copyrightyear{2020}
\acmYear{2020}
\setcopyright{acmcopyright}\acmConference[MM '20]{Proceedings of the 28th ACM International Conference on Multimedia}{October 12--16, 2020}{Seattle, WA, USA}
\acmBooktitle{Proceedings of the 28th ACM International Conference on Multimedia (MM '20), October 12--16, 2020, Seattle, WA, USA}
\acmPrice{15.00}
\acmDOI{10.1145/3394171.3413694}
\acmISBN{978-1-4503-7988-5/20/10}

\begin{document}
\fancyhead{}

\title{Self-supervised Video Representation Learning \\ Using Inter-intra Contrastive Framework}

\author{Li Tao}
\affiliation{%
 \institution{The University of Tokyo}
}
\email{taoli@hal.t.u-tokyo.ac.jp}

\author{Xueting Wang}
\authornote{Corresponding Author}
\affiliation{%
 \institution{The University of Tokyo}
}
\email{xt_wang@hal.t.u-tokyo.ac.jp}

\author{Toshihiko Yamasaki}
\affiliation{%
 \institution{The University of Tokyo}
}
\email{yamasaki@hal.t.u-tokyo.ac.jp}

\renewcommand{\shortauthors}{Tao, Wang and Yamasaki.}

\begin{abstract}
We propose a self-supervised method to learn feature representations from videos. A standard approach in traditional self-supervised methods uses positive-negative data pairs to train with contrastive learning strategy. In such a case, different modalities of the same video are treated as positives and video clips from a different video are treated as negatives. Because the spatio-temporal information is important for video representation, we extend the negative samples by introducing intra-negative samples, which are transformed from the same anchor video by breaking temporal relations in video clips. With the proposed Inter-Intra Contrastive (IIC) framework, we can train spatio-temporal convolutional networks to learn video representations. There are many flexible options in our IIC framework and we conduct experiments by using several different configurations. Evaluations are conducted on video retrieval and video recognition tasks using the learned video representation. Our proposed IIC outperforms current state-of-the-art results by a large margin, such as 16.7\% and 9.5\% points improvements in top-1 accuracy on UCF101 and HMDB51 datasets for video retrieval, respectively. For video recognition, improvements can also be obtained on these two benchmark datasets. 
Code is available at \url{https://github.com/BestJuly/Inter-intra-video-contrastive-learning}.
\end{abstract}




\maketitle

\section{Introduction}
There are many video understanding tasks, such as video captioning and video segmentation. These tasks rely on effective motion representation extractors, which are usually trained on the basis of video recognition. For video recognition, the works presented in \cite{wang2016temporal,c3d,res3d,r3d,s3d,i3d} have explored different network architectures. In~\cite{feichtenhofer2016spatiotemporal,simonyan2014two,feichtenhofer2016convolutional}, an additional optical flow stream was used to form a two-stream model. With optical flow, better results were achieved~\cite{i3d,r3d,s3d}. 
Hara et al.~\cite{res3d} argued that they can imitate image recognition procedures, which means that the performance can be significantly improved with large datasets. 

\begin{figure}[t]
  \centering
  \includegraphics[width=220pt]{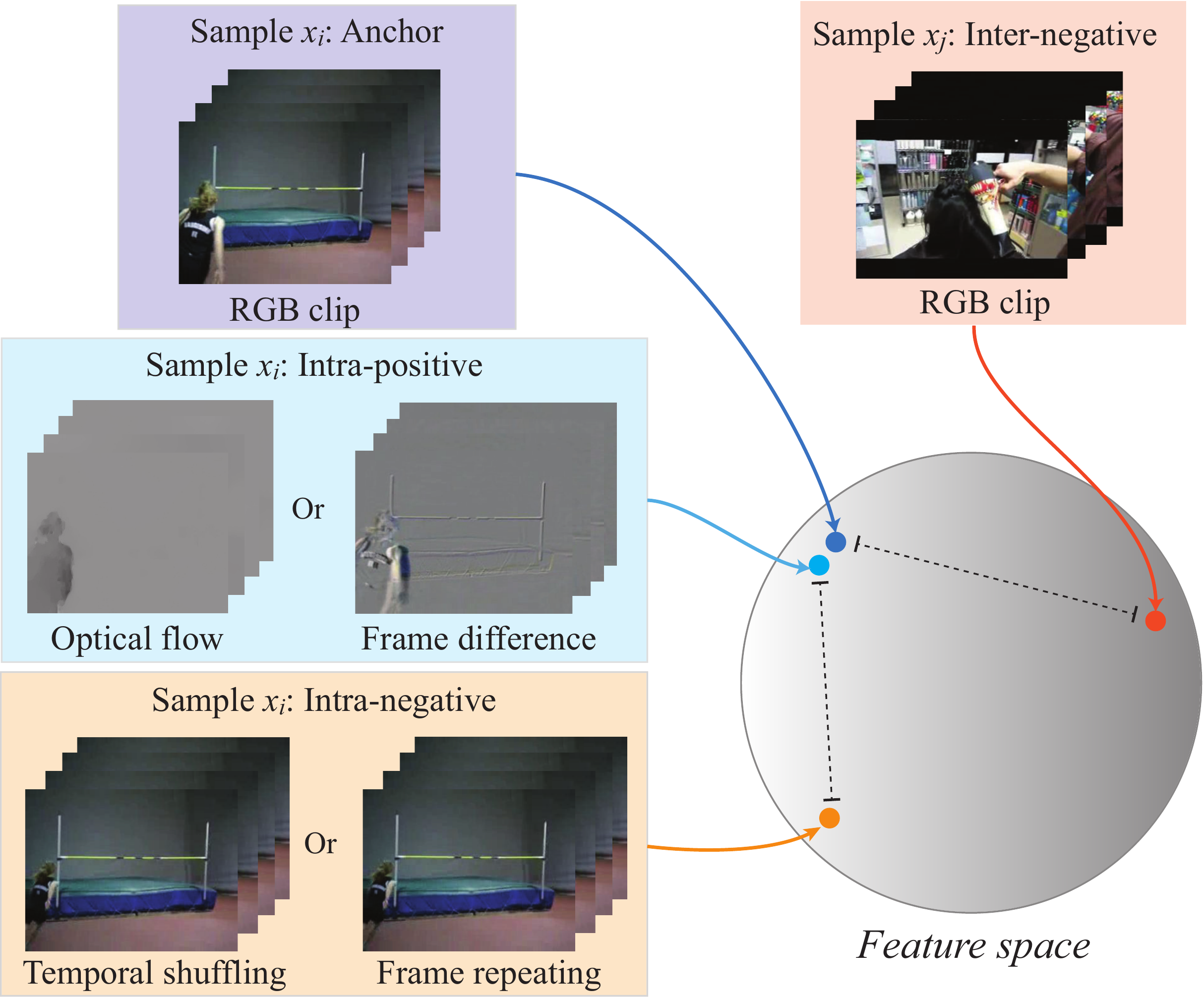}
  \caption{General idea of IIC. Given video $x_i$, different views of this video are treated as positives, and those features are constrained to be close to each other. Data from other videos are treated as negatives. Temporal relations in the anchor view will be broken down to generate intra-negative samples, which are also treated as negatives to help the model learn temporal information.}
  \label{general}
\end{figure} 

Though larger datasets are helpful for video understanding tasks and numerous unlabeled videos are available on the Internet, annotating new video datasets requires a wealth of resources. Ensuring the performance of training action classification networks usually requires properly trimmed action video clips, which makes the situation more serious. Therefore, it is valuable if the unlabeled videos can be leveraged to facilitate learning. From this point of view, self-supervised learning is drawing a lot of attention these days beacuse it does not require any labels. 

Many self-supervised learning techniques are proposed for image data. There are several designed tasks such as solving jigsaw puzzles~\cite{noroozi2016unsupervised}, image inpainting~\cite{pathak2016context}, and image color channel prediction~\cite{zhang2016colorful}. For video data, existing works~\cite{misra2016shuffle,fernando2017self,lee2017unsupervised,xu2019self} have focused on changing the temporal information and making models sensitive to the differences. The aforementioned methods can be classified into a single category, which we call as \textbf{intra-sample learning} because all operations are carried out in the sample itself. For example, if a video contains several frames, shuffling frames to change their order is performed within this sample.

In addition to intra-sample learning, \textbf{inter-sample learning} is also a kind of self-supervised learning technique. For image data, when we have an anchor image crop, crops from the same image are treated as positives while crops from different images are treated as negatives. If a model can distinguish whether samples come from the same sample set or not, it is certain that this model can extract discriminative features, which may be good feature representations. The procedure is almost the same for video data.

For intra-sample learning methods, tasks should be carefully designed, whereas inter-sample learning methods are simpler. However, for inter-sample learning, whether good temporal information can be extracted relies on the model itself. Further, if spatial information is sufficient enough as compared to its temporal information, the model will not be helpful for other video related tasks. Therefore, our goal is to learn better representations that can capture rich temporal information. To do so, we break the temporal relationship of the anchor sample to generate intra-negative samples. Then the models can learn spatial differences as well as temporal differences between samples. In particular, we adapt the recently proposed method of \textit{Contrastive Multiview Coding} (CMC)~\cite{tian2019contrastive}, extend it with more generalized video recognition models, and improve it by introducing intra-negative sample learning. The general idea of our proposed method is illustrated in Fig.~\ref{general}

In this paper, we propose Inter-Intra Contrastive (IIC) learning framework in videos, which are built on the basis of many existing techniques such as inter-intra learning, contrastive learning, and deep representation learning. To the best of our knowledge, we are the first to focus and apply these techniques together to videos. Recent self-supervised learning in videos has mainly used intra-sample learning methodologies, and video retrieval and video recognition tasks were considered as evaluation tasks. Our main contribution is to combine the advantages of inter- and intra-sample learning and establish a general framework for self-supervised learning for video representation. We have explored several options towards best practices in our framework. In addition, for both video retrieval and video recognition tasks, we outperform the existing state-of-the-art results by a notably large margin. 

Our contributions are summarized as follows:
\begin{itemize}
\item We generate intra-negative samples by breaking temporal relations, which encourage the model to learn rich temporal information as well as spatial information, and it is helpful for motion feature representation.
\item We extend the contrastive multiview coding framework to have an inter-intra style for video representation, where many useful options in the framework are also provided.
\item Our experiments show that by using the proposed IIC framework, significant improvements over the state-of-the-art methods are achieved with the same network architecture.
\end{itemize}

\section{Related works}
In this section, we divide the existing self-supervised learning methods into two categories according to their learning style, namely inter-sample learning and intra-sample learning. In addition, because we focus on video representation, we add another subsection to briefly introduce the techniques in video understanding.

\subsection{Intra-sample learning}
For intra-sample learning methods, the constraints are in the sample itself. By using different transformation functions, some relations are broken down even though statistical or semantic information remains. Different target tasks are carefully designed to help train the model.

Self-supervised learning methods are close to unsupervised representation learning, and include methods such as autoencoders~\cite{hinton2006reducing} and variational autoencoders~\cite{kingma2013auto}.
Noroozi et al.~\cite{noroozi2016unsupervised} proposed to learn features by solving Jigsaw puzzles. Pathak et al.~\cite{pathak2016context} trained context inpainting models to learn feature representation. Gidaris et al.~\cite{gidaris2018unsupervised} rotated images and trained models by predicting the rotated angles.

Because videos have an additional temporal axis compared to images, for video representation learning, how to efficiently extract temporal information is important. There are many existing works focusing on temporal orders~\cite{misra2016shuffle,fernando2017self,lee2017unsupervised,xu2019self}. Misra et al.~\cite{misra2016shuffle} treated several video frames as a sequence, and trained a network to distinguish whether these video frames were in the right order. Odd-one-out network (O3N)~\cite{fernando2017self} was proposed to identify unrelated or odd video clips. Order prediction network (OPN)~\cite{lee2017unsupervised} shuffled frames and trained networks to predict the correct order of input frames. Similar to OPN, Xu et al.~\cite{xu2019self} set video clips as inputs and used 3D convolutional networks to predict the order. In addition to focusing on the temporal order, Wang et al.~\cite{wang2019self} proposed regressing motion and appearance statistics to learn video representations. Kim et al.~\cite{kim2019self} proposed training models by completing space-time cubic puzzles. Luo et al.~\cite{luo2020video} applied one transformation from several options, including spatial rotation and temporal shuffling, to video clips and trained models to recognize which action has been applied. The performance of these methods depends highly on the special designed tasks.

\subsection{Inter-sample learning}
For inter-sample learning methods, features from the same sample should be close to each other while the distance between different samples should be far from each other. 

In~\cite{sermanet2018time}, frames from the same video were treated as positives while frames from different videos were negatives. And triplet loss~\cite{hermans2017defense} was used to train the network. After contrastive losses~\cite{hadsell2006dimensionality} were proposed, contrastive learning has become the core of self-supervised learning, especially on image data. Contrastive Predictive Coding~(CPC)~\cite{oord2018representation} used sequential data to learn the future from the past. Deep InfoMax~\cite{hjelm2018learning} and Instance Discrimination~\cite{wu2018unsupervised} learned to maximize information probability from the same sample. Contrastive Multiview Coding~(CMC)~\cite{tian2019contrastive} used different views for the same sample and minimized the distance between different views while maximizing the distance between different samples. 
MoCo~\cite{he2019momentum} used a momentum encoder with a momentum-updated encoder to conduct contrastive learning. In SimCLR~\cite{chen2020simple}, different combinations of data augmentation methods were experimented for paired samples. Note that none of these inter-sample learning methods require specially designed tasks.

Our proposed method is closely related to CMC~\cite{tian2019contrastive}, where multi-view coding is used. Most methods treat data from different samples as negatives. In our research, we deal with video data and extend negative samples by breaking the temporal relationships in video clips.

\begin{figure*}[t]
  \centering
  \includegraphics[width=440pt]{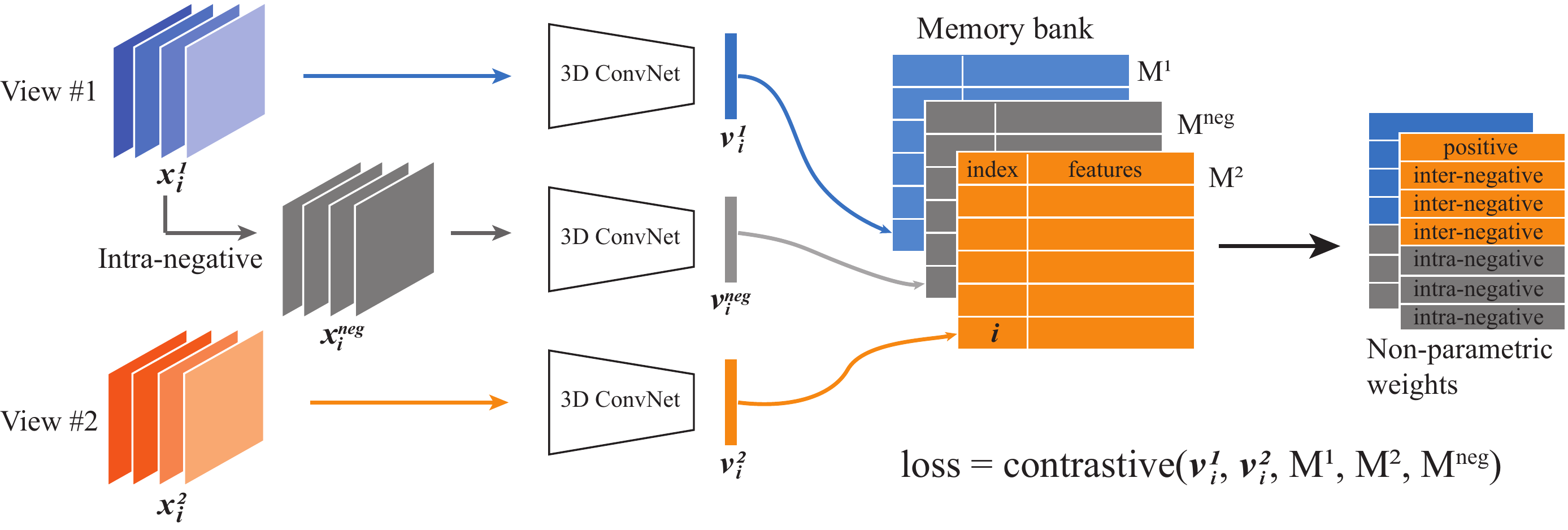}
  \caption{The main framework of IIC. Intra-negative samples are generated from the first view by breaking its temporal relationship. Video clips from two different views as well as the intra-negative clip are used in one iteration. Features are processed with three corresponding memory banks and non-parametric weights are obtained. Contrastive loss is used for the optimization of the network.}
  \label{framework}
\end{figure*} 

\subsection{Video representation}
Previous self-supervised learning methods have mainly been applied on images. Some video representation learning methods still use image frames as inputs~\cite{misra2016shuffle,fernando2017self,lee2017unsupervised}, which do not enjoy the benefits from new techniques related to video understanding.

For video representation, many supervised methods have been proposed. Temporal Segment Networks (TSN)~\cite{wang2016temporal} split one video into several segments and sampled one frame from each segment as the input data of a 2D CNN. In addition to a single 2D CNN for RGB data, two-stream ConvNets~\cite{simonyan2014two, feichtenhofer2016convolutional, feichtenhofer2016spatiotemporal} have been used with an additional optical flow stream. Recently, spatio-temporal convolution (3D-CNN) was applied to video recognition task. In C3D~\cite{c3d}, Tran et al. used 3D convolutional layers to form their network and achieved good performance. 3D convolutional versions of ResNet~\cite{resnet} and Inception net~\cite{inception}, R3D~\cite{res3d} and I3D~\cite{i3d}, were proposed and showed promising performance on benchmark datasets~\cite{ucf101,hmdb,kinetics}. By separating one 3D convolutional kernel into two steps, a spatial part and a temporal part, R(2+1)D~\cite{r3d} and S3D~\cite{s3d} were proposed. Those trained models were proved to be effective feature extractors when applied to other video related tasks.

The aforementioned models can also be used in self-supervised learning to handle video data. By replacing a 2D CNN with a 3D CNN, \cite{xu2019self} reported better performance than \cite{lee2017unsupervised} as their target tasks were the same, predicting the temporal orders of inputs. C3D, R3D, and R(2+1)D were used in~\cite{xu2019self,luo2020video} and proved to be effective for self-supervised learning with video data. Similar to these methods, 3D convolutional networks are used in our proposed framework. 

\section{Methods}
Our goal is to learn discriminative feature representations from videos, not only for distinguishing one action from another, but also for capturing rich temporal information. The entire IIC framework is shown in Fig.~\ref{framework}. In this section, we start from the novel input part, and then elaborate on contrastive learning with these inputs. Because we simplify the model by using only one network to cope with three kinds of input data, an unique joint retrieval method will also be introduced.

\subsection{Multi-view and intra-negative inputs}
\label{intraneg}
We denote video data from two different views as $X^1$ and $X^2$, and data with the same video id $i$ from these two views as $x_i^1$ and $x_i^2$, respectively. The definition of view here is broad, including data in different color space, depth information, segmentation information, etc.  Without loss of generality, only two views are used in this work, which can be extended with more views of the same sample. We use a 3D convolutional network as our backbone. Therefore, the referred data $x_i^1$ and $x_i^2$ are in shape $THWC$, where $T$ continuous frames with height $H$ and width $W$ are stacked together. $C$ is the channel number of frames. Temporal information relies on the connections among $T$ stacked frames.

For multi-view contrastive learning, feature $v_i^1$ and feature $v_i^2$ should be close to each other because those features are extracted from the same video $i$. In addition, feature $v_i^1$ should be far from features $v_j^1~({\rm for}~j \neq i)$. This is effective enough for images. On the other hand, video data have one more dimension. When the same person behaves in opposite ways, e.g. \textit{standing up} and \textit{sitting down}, the appearance information of each frame is similar, however, traditional contrastive learning methods will easily be fooled. 

\begin{figure}[t]
  \centering
  \includegraphics[width=180pt]{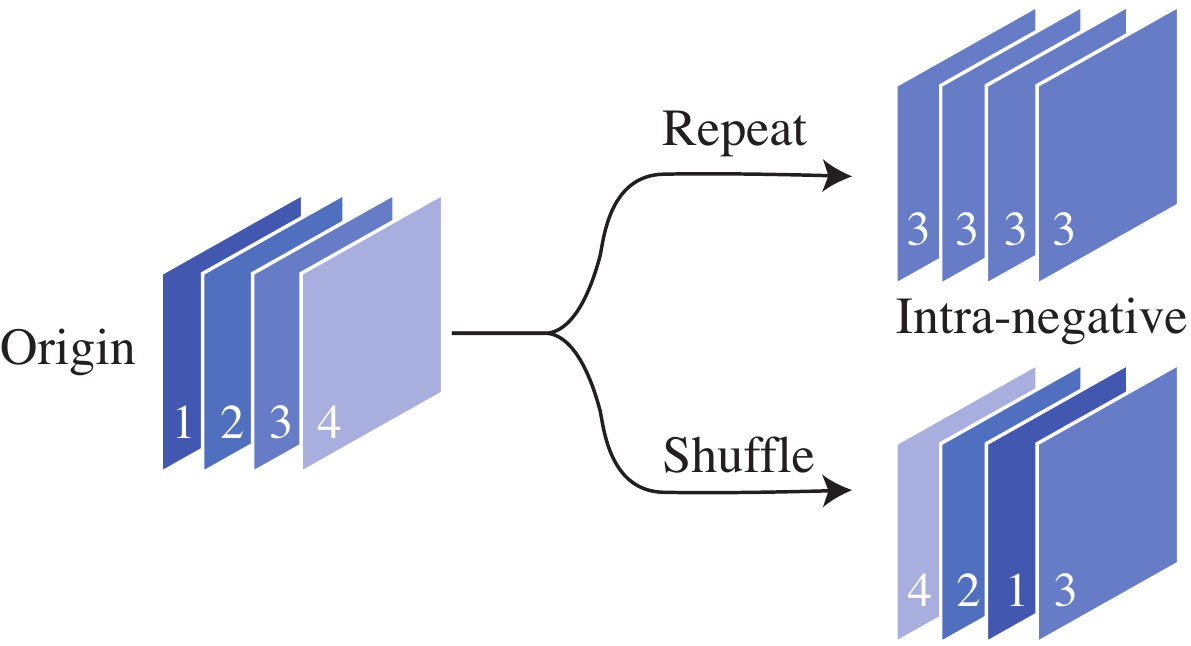}
  \caption{Generating intra-negative samples from original video clips.}
  \label{generate_intra}
\end{figure}

Here, we introduce intra-negative samples in multi-view contrastive learning for videos by breaking the temporal relationship. For one video clip, the data $x_i^1$ is a set of frames. To simplify, we use $\{frame_1, ..., frame_T\}$ to represent a set of temporally-ordered frames. Two kinds of methods, frame repeating and temporal shuffling, are proposed to break the temporal relationship of a video clip (Fig.~\ref{generate_intra}).

\noindent{\bf Frame repeating.} One frame that is randomly selected from one video clip is repeated $T$ times to generate intra-negative samples (eq.~\ref{repeat}). Then all frames in this video clip are the same and no movements exist anymore, despite the scene of this intra-negative sample is almost the same as the source.

\begin{equation}
  \label{repeat}
  x_{repeat} = \{frame_k, ..., frame_k\},~k = random(1, T). \\
\end{equation}

\noindent{\bf Temporal shuffling.} In the original video clip, frames are in the correct sequence. By randomly shuffling the frames (eq.~\ref{shuffle}), the actions will be strange and the corresponding action information should be different. After this transformation, the global statistical information of the intra-negative sample remains the same as its source.

\begin{equation}
  \label{shuffle}
  x_{shuffle} = shuffle(x),~where\ x_{shuffle} \neq x.
\end{equation}

Note that intra-negative samples can be generated from data for both view 1 and view 2, and both generating functions can be used simultaneously. To simplify, in this paper, we only generated intra-negative samples from view 1 (original RGB video clips) and only one generating function was used for each experiment. Then $x^{neg}$ is used to represent intra-negative samples from $x^1$.

\subsection{Contrastive learning}
Similar to learning with triplet loss, which aims to learn an embedding that separates the negative samples from the positive and the anchor, contrastive learning aims to separate samples from two different distributions. In traditional multi-view coding~\cite{tian2019contrastive}, the sample pairs $\{x_i^1, x_i^2\}$ are positives while $\{x_i^1, x_j^2\} (i\neq j)$ are negatives. Because we generate intra-negative samples, the negatives are extended by adding pairs $\{x_i^1, x_j^{neg}\}$, where $j$ can be equal to $i$.

A discriminative function $h_\theta(\cdot)$ is used to ensure that positive pairs have high values while the value for negative pairs should be low. The function is trained by selecting a single positive sample from a set of data. After feature $v_i^1$ has been extracted, traditional contrastive learning methods train this function to correctly select a positive sample out of a set $S=\{v_1^2, ..., v_i^2, ..., v_{k+1}^2\}$, which contains one positive sample $v_i^2$ and $k$ negative samples. In our proposed method, another set $S^{neg}=\{v_1^{neg}, ..., v_{k+1}^{neg}\}$ is also used which only contains negative samples. The loss function is similar to recent works for contrastive learning~\cite{tian2019contrastive, oord2018representation, gutmann2010noise}:

\begin{equation}
  \label{loss_1}
  \mathcal{L}_{contrast}^{v_i^1} = - \log\frac{h_\theta(\{v_i^1, v_i^2\})}{\sum_{j=1}^{k+1}{h_\theta(\{v_i^1, v_j^2\})} + \sum_{j=1}^{k+1}{h_\theta(\{v_i^1, v_j^{neg}\})}}.
\end{equation}
Here, $k$ is the number of negative samples, which can be equal to $N-1$, where $N$ is the total number of training samples. To accelerate training, we randomly select $k$ samples from $N$ where $k \ll N$.

The critic $h_\theta(\cdot)$ is implemented by feature representations using the non-parametric softmax technique~\cite{wu2018unsupervised}. Then we can compute this function as the following:
\begin{equation}
  h_\theta(\{v_i^1, v_j^2\}) = \exp\left(\frac{v_i^1\cdot v_j^2}{\Arrowvert{v_i^1}\Arrowvert\cdot \Arrowvert{v_j^2}\Arrowvert} \cdot \frac{1}{\tau}\right),
\end{equation}
where $\tau$ is a hyper-parameter that controls the range of the results. In practice, three memory banks are used to store the extracted features from previous iteration, and these features function as weights in the non-parametric softmax learning~\cite{wu2018unsupervised}.

Eq.~\ref{loss_1} only treats view~1 as an anchor. When treating view~2 as an anchor, symmetrically, another loss can be calculated and they are added to form the final loss function:
\begin{equation}
  \mathcal{L} = \mathcal{L}_{contrast}^{v^1} + \mathcal{L}_{contrast}^{v^2}.
\end{equation}

To summarize this section, we write the process flow of our proposal in Appendix~\ref{alg}.

\subsection{Joint representation}
For learning video representations with supervision, different modality data require different models because the input channels are usually different. Because stacked frame differences, which have the same shape as the original RGB video clips, have succeeded in supervised learning~\cite{tao2020rethinking}, it is possible to use one network to handle video data in different views. In practice, we constrain data from different views in the same shape and use only one model to process data from different views. The options for different views are original RGB clips, optical flow ($u$ or $v$) frame clips and stacked frame differences, and we used RGB clips as the anchor view. In the following part, for convenience, we use \textbf{residual clip} to represent stacked frame differences.

\begin{figure}[t]
  \centering
  \includegraphics[width=200pt]{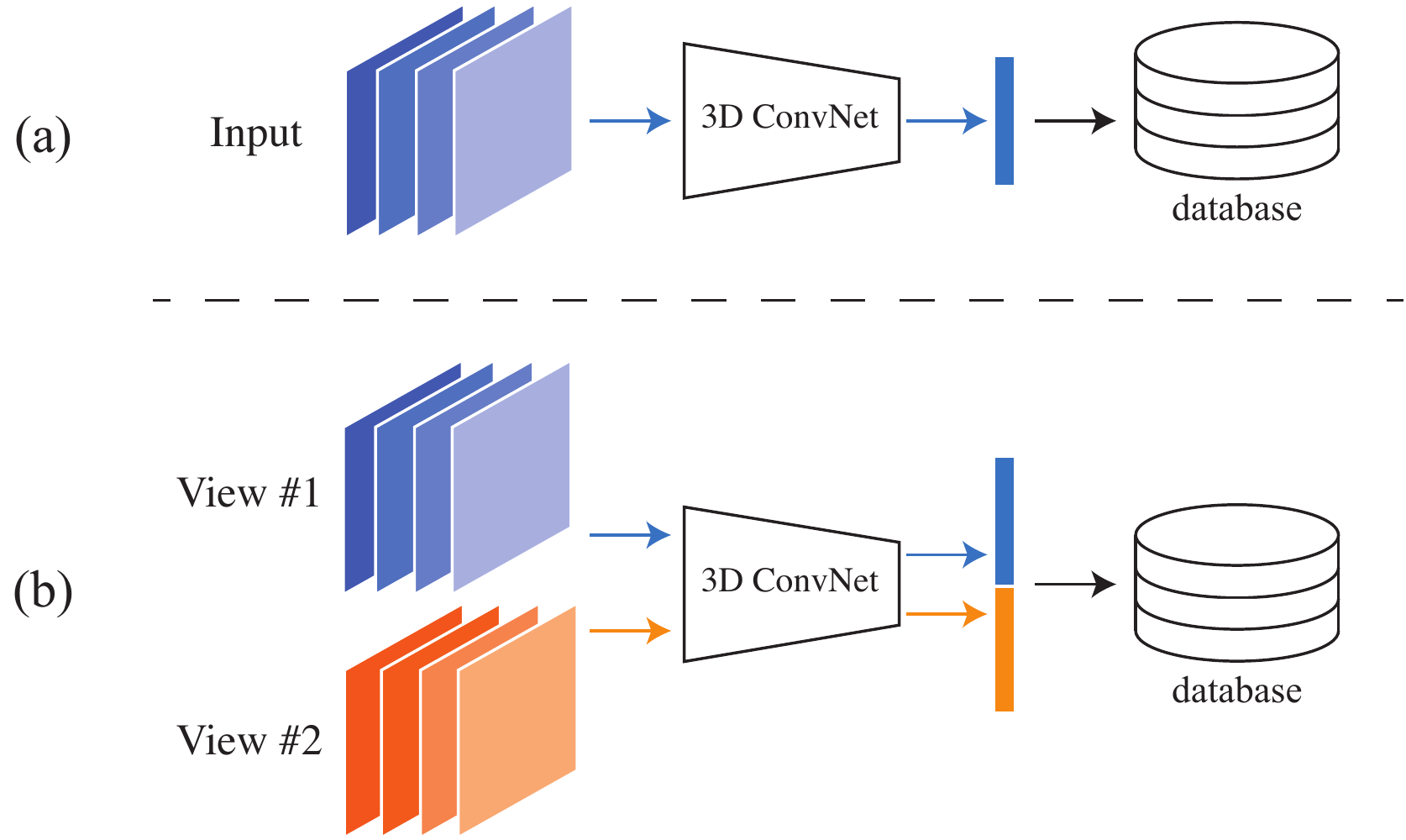}
  \caption{Joint retrieval using two different kinds of input data and only one network.}
  \label{joint_retrieval}
\end{figure} 

After training is complete, the model can be used to extract features from different views of videos. These features can be concatenated to form two-view features and can be applied to video retrieval (Fig.~\ref{joint_retrieval}). We address this because for two-stream methods, features extracted from \textbf{different} models can be also concatenated to represent one video. With our multi-view contrastive learning approach, different views of the same video are set as the input of only \textbf{one} model. According to our experiments, it is sufficient to use one model to handle different view data. This strategy can help to extract more comprehensive representations of videos.

\section{Experiments}
We conducted extensive experiments to evaluate our proposed IIC framework and its transferability when applied to other tasks or datasets. Because there are several options in our framework, we first elaborate on some option configurations on video retrieval tasks because this can be evaluated directly when the training has been performed. The trained models are also treated as one kind of weight initialization strategy and by fine-tuning trained models on video recognition datasets, we can further evaluate the performance of our model.

\subsection{Datasets}
There are several existing labeled benchmark datasets in video recognition: UCF101~\cite{ucf101}, HMDB51~\cite{hmdb}, something-something~\cite{goyal2017something}, and Kinetics400~\cite{kinetics}. For fair comparison, we follow~\cite{xu2019self,tian2019contrastive,luo2020video} and use the UCF101 and HMDB51 datasets. The UCF101 dataset consists of 13,320 videos in 101 action categories. HMDB51 is comprised of 7,000 videos with a total of 51 action classes. If not specially declared, experiments were on \textit{split}~1 for both UCF101 and HMDB51 datasets.

\subsection{Evaluation tasks}
Our goal is to learn effective and discriminative video representations using IIC learning framework. After training has been performed, the direct way to evaluate is to use the trained model to extract video features, then video retrieval can be tested easily. UCF101 and HMDB51 are two different datasets. We trained our model only on UCF101 \textit{split}~1 and performed video retrieval on both UCF101 and HMDB51 datasets. When applied to HMDB51, the model generalization ability was mainly tested. To evaluate whether good feature representations are learned with our proposed method, we also conducted experiments by fine-tuning trained models on both UCF101 and HMDB51 datasets.

\subsection{Options in our IIC framework}
\noindent{\bf Multiple Views.} 
For video representation learning, traditional RGB input and the corresponding optical flow were set as two common views~\cite{sayed2018cross, tian2019contrastive}. Optical flow data require additional calculation and storage. In addition to optical flow, in video tasks, frame difference has also been used in existing works~\cite{wang2016temporal} with 2D ConvNets. Residual frames with 3D ConvNets have been proved to be more effective compared to original RGB video clips~\cite{tao2020rethinking}, which can also be set as one view of video data. In our experiments, we chose RGB video clips and another view from residual frames or optical flow. Then the modality for the second view was from ${res, u}$, and ${v}$.

\noindent{\bf Backbone networks.} 3D convolutional kernels have been proved effective in many recent works. Recent self-supervised learning methods~\cite{xu2019self, luo2020video} used C3D~\cite{c3d}, R3D~\cite{res3d}, and R(2+1)D~\cite{r3d} as their network backbones. We mainly used R3D in our experiments, where each residual block consists of two 3D convolution layers.

\noindent{\bf Intra-negative generation.} As we discussed in section~\ref{intraneg}, we have two ways to generate intra-negative samples, frame repeating and temporal shuffling. In our experiments, both were tested and only one generation method was used in each experiment.

\subsection{Implementation details}
The input preparation part follows~\cite{c3d}. Sixteen successive frames are sampled with size $128\times171$ to form a video clip for both view 1 and view 2. Random spatial cropping was conducted to generate an input data of size $16\times112\times112$, where the channel number $3$ was ignored. For the data from the second view, when residual frames were used, we shifted the RGB video clip along the temporal axis and calculated the difference between the original clip and the shifted clip. When optical flow was used, because traditional tv-l1 optical flow calculates motion features in two directions--$u$ and $v$, we picked one direction and duplicate it to generate optical flow clips with channel dimension $3$. Then we can use one 3D ConvNet to handle data from different views.

When performing temporal shuffling, similar to~\cite{luo2020video}, one clip was divided into four sub-clips, and we shuffled the sub-clips to conduct temporal shuffling. 

When training unsupervised procedure, the batch size is set to 16 and training lasted for 240 epochs. The initial learning rate was set to 0.01 and it was updated by multiplying a fixed rate of 0.1 after 45, 90, 125 and 160 epochs. In non-parametric learning part, $2k$ negative samples were sampled from memory banks, with $k$ set to 1,024 for all our experiments. Video retrieval was conducted by K nearest neighbor search. When evaluating task transferability, we used our trained models as an initialization strategy and the learning rate was set to 0.001 for fine-tuning. The best performance on the validation dataset was used for testing.

\section{Results and discussion}
In this section, we first report our ablation studies with several option configurations that were mentioned above. Then, we outline the four option configurations that were selected to compete with the state-of-the-art methods in self-supervised spatio-temporal learning.

\subsection{Ablation study}
\noindent{\bf Joint retrieval.} 
Because the models were evaluated on video retrieval task after self-supervised training process was done, we first conducted ablation studies on the effectiveness of proposed joint retrieval strategy. Results are shown in Table~\ref{table:joint}. Here, the performance was tested by treating residual frames as the second view. Experiments were conducted on UCF101 \textit{split}~1.

\begin{table}[tb]
\centering
  \caption{Effectiveness of joint retrieval.}
  \setlength{\tabcolsep}{1mm}{
  \begin{tabular}{c c c c c c c}
      \hline 
      intra-neg & retrieval mode & top1 & top5 & top10 & top20 & top50 \\
      \hline
      $\times$ & view1: rgb & 32.5 & 47.6 & 57.2 & 68.3 & 81.0 \\
      $\times$ & view2: res & 32.4 & 50.6 & 60.4 & 69.8 & 80.6 \\
      $\times$ & joint & \textbf{34.6} & \textbf{52.1} & \textbf{61.8} & \textbf{71.4} & \textbf{82.4} \\
      \hline
      \checkmark & view1: rgb & 34.8 & 51.6 & 60.8 & 69.7 & 80.6 \\
      \checkmark & view2: res & 33.4 & 53.2 & 62.5 & 72.0 & \textbf{83.6} \\
      \checkmark & joint & \textbf{36.5} & \textbf{54.1} & \textbf{62.9} & \textbf{72.4} & 83.4 \\
      \hline
  \end{tabular}}
  \label{table:joint}
\end{table}

As we can see from the table, by using \textbf{only one model} to process data from two different views, when features were concatenated to represent videos, nearly 2\% points improvements could be achieved at top-1 retrieval accuracy. This indicates that these features were more robust for video representation. We set the proposed joint retrieval strategy as the default setting for the following experiments. 
\\

\noindent{\bf Option configurations.} 
We conducted experiments in three parts: 1. whether to use intra-negative samples or not; 2. which intra-negative sample generation method to use; 3. which modality was to be chosen for the second view. 

\begin{table}[tb]
\centering
  \caption{Ablation studies on different option configurations. The best performances are in \textbf{bold} for each \textit{modality}. Note that RGB video clips are fixed for view 1 and the modalities here represent only for view 2.}
  \scalebox{0.95}{
  \setlength{\tabcolsep}{0.3mm}{
  \begin{tabular}{c c c c c c c c}
      \hline
      Intra-neg & type & view2 & top1 (\%) & top5 (\%) & top10 (\%) & top20 (\%) & top50 (\%)\\
      \hline
      $\times$ & -          & res & 34.6 & 52.1 & 61.8 & 71.4 & 82.4 \\
      $\times$ & -          & u   & 37.5 & 54.8 & 64.1 & 72.6 & 83.5 \\
      $\times$ & -          & v   & 34.7 & 53.4 & 63.5 & 72.0 & 82.9 \\
      \hline
      \checkmark & repeat   & res & \textbf{36.5} & \textbf{54.1} & \textbf{62.9} & \textbf{72.4} & \textbf{83.4} \\
      \checkmark & repeat   & u   & \textbf{41.8} & \textbf{60.4} & \textbf{69.5} & \textbf{78.4} & \textbf{87.7} \\%
      \checkmark & repeat   & v   & 34.3 & 55.9 & 65.3 & 73.2 & 83.3 \\
      \hline
      \checkmark & shuffle  & res & 34.6 & 53.0 & 62.3 & 71.7 & 82.4 \\
      \checkmark & shuffle  & u   & 39.2 & 57.7 & 66.6 & 75.1 & 84.7 \\
      \checkmark & shuffle  & v   & \textbf{42.4} & \textbf{60.9} & \textbf{69.2} & \textbf{77.1} & \textbf{86.5} \\
      \hline
  \end{tabular}}}
  \label{table:ablation}
\end{table}

As we can see from Table~\ref{table:ablation}, no matter which modality was used as the second view, all retrieval accuracies were better when the proposed intra-negative samples were used in multi-view coding. 

For residual frames, which can be treated as one solution without optical flow, the best performances were achieved by using frame repeating strategy. If temporal shuffling was used, improvements could also be obtained.

When optical flow data was used, the results outperformed those using residual frames as the second view. This makes sense because optical flow data requires additional calculations and is designed to represent motion features, which is suitable for video representation. The best performance on the top-1 and top-5 accuracies was achieved by using temporal shuffling to generate intra-negative samples with $v$ data while for the top-10, top-20, and top-50 accuracies, using frame shuffling with $u$ data was better. This is reasonable because $u$ and $v$ are two dimensions of optical flow data, which record movements in two directions. The performance relies on the main movement directions in videos. This also highlighted one limitation of the current settings, and indicated that it may be better to use both $u$ and $v$ data to form the optical flow stream for self-supervised learning, which requires two different models. In the present study, we used only one model for data from both views. We intend to explore the use of multiple models for different views in our future work.

All the aforementioned results indicate that by introducing intra-negative samples as additional negatives in contrastive learning, the model could focus more on learning discriminative temporal information, which is helpful for feature learning.

\subsection{Visualization: feature embedding}
\begin{figure}[t]
  \centering
  \includegraphics[width=200pt]{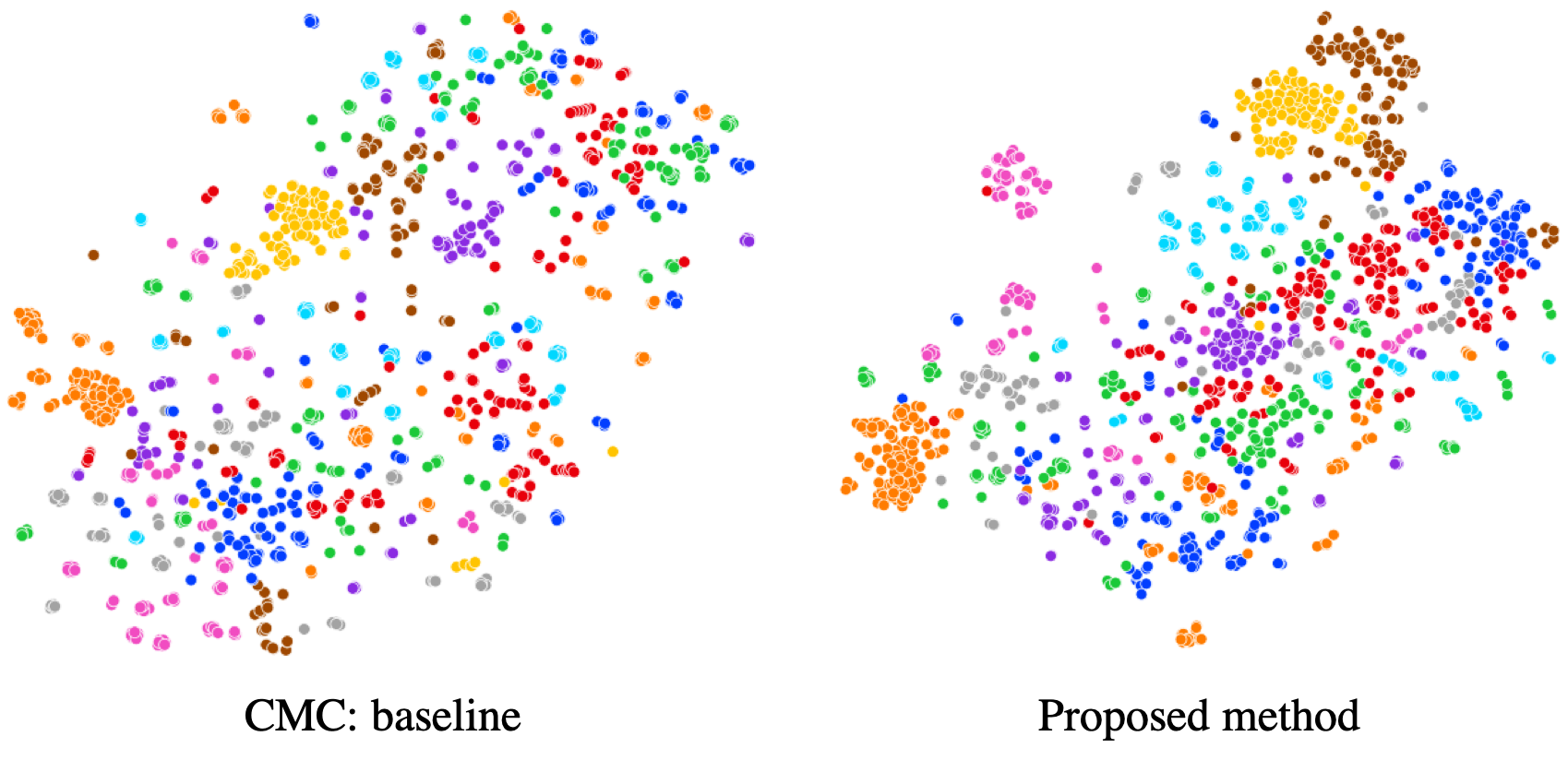}
  \caption{Feature visualization by t-SNE. Features extracted by our proposed method are more semantically separable compared to the baseline, which does not use intra-negative samples during training. Each video is visualized as a point, with videos belonging to the same action category having the same color.}
  \label{tsne}
\end{figure} 

Before applying our proposed method to other evaluation tasks, we used trained models to extract video features and qualitatively evaluated these features by visualization in order to verify whether good feature representations have been learned. We selected videos from UCF101 which belong to the first 10 categories (arranged by action names in alphabetical order). Features were projected to 2-dimensional space using t-SNE~\cite{maaten2008visualizing}. Fig.~\ref{tsne} visualizes the embedding of the features extracted by the baseline~\cite{tian2019contrastive} and our proposed method. It is obvious that with intra-negative samples, the trained models showed better clustering ability for video data. We quantitatively observe that better video representations were able to be learned by our proposed method.

\begin{figure*}[t]
  \centering
  \includegraphics[width=440pt]{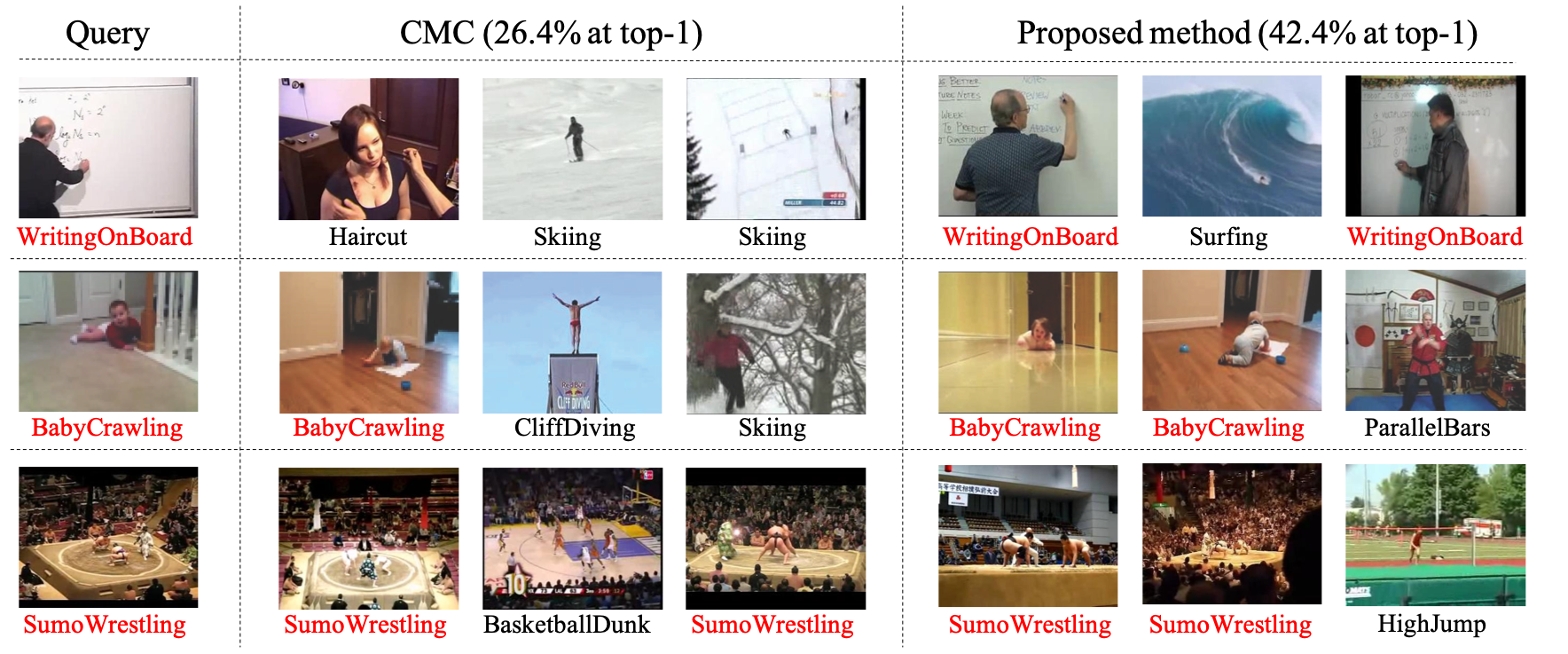}
  \caption{Comparison of video retrieval results with baseline method. Red fonts indicate correct retrieval results.}
  \label{retrieval}
\end{figure*} 

\subsection{Comparison: video retrieval}

For a fair comparison, we trained our models on UCF101 and tested them on both UCF101 and HMDB51 datasets. Specially, the most related work CMC~\cite{tian2019contrastive} mainly focused on Image-related tasks and for video representation part, two CaffeNets~\cite{krizhevsky2012imagenet} with two views--RGB and optical flow--were used. We reimplemented their work using a more common model, ResNet-18~\cite{resnet}. The model is a 2D convolutional style and for both RGB and optical flow data, the spatial size is $224\times224$. In addition, we experimented with a one-view model to prove the effectiveness of multiple views for videos. To simplify, we set the modality of the second view to be RGB which is the same as the first view.

\begin{table}[t]
  \setlength{\tabcolsep}{0.8mm}{
  \centering{
  \caption{Video retrieval performance on UCF101.}\label{table:retrieve_ucf101}
  \scalebox{0.95}{
  \begin{tabular}{lccccc}
  \hline
  Methods&top1 (\%)&top5 (\%)&top10 (\%)&top20 (\%)&top50 (\%)  \\
  \hline
  Jigsaw~\cite{noroozi2016unsupervised} &19.7&28.5&33.5&40.0&49.4 \\
  OPN~\cite{lee2017unsupervised} &19.9&28.7&34.0&40.6&51.6 \\
  B\"uchler~\cite{buchler2018improving} &25.7&36.2&42.2&49.2&59.5 \\
  \hline
  R3D (random) &9.9&18.9&26.0&35.5&51.9 \\
  VCOP~\cite{xu2019self} &14.1 & 30.3&40.4&51.1&66.5\\
  VCP~\cite{luo2020video}&18.6&33.6&42.5&53.5&68.1 \\
  \hline
  One-view & 26.2 & 39.3 & 46.8 & 55.6 & 66.8 \\ 
  CMC~\cite{tian2019contrastive} & 26.4 & 37.7 & 45.1 & 53.2 & 66.3 \\
  \hline
  IIC (repeat + res) & \textbf{36.5} & \textbf{54.1} & \textbf{62.9} & \textbf{72.4} & \textbf{83.4} \\
  IIC (repeat + u) & 41.8 & 60.4 & \textbf{69.5} & \textbf{78.4} & \textbf{87.7} \\
  IIC (shuffle + res) & 34.6 & 53.0 & 62.3 & 71.7 & 82.4 \\
  IIC (shuffle + v) & \textbf{42.4} & \textbf{60.9} & 69.2 & 77.1 & 86.5 \\
  \hline
  \end{tabular}}}
  }

\end{table}

The results on the UCF101 dataset are shown in Table~\ref{table:retrieve_ucf101}. The compared models mostly design a special task for the model to learn, which belong to the intra-sample learning category. IIC treats every different samples as negative. In addition, intra-negative samples are generated to enable the model to learn more temporal clues. We picked four option configurations, two of which do not use optical flow data. As shown in this table, when only RGB video clips were used with contrastive learning, the top-1 accuracy was already higher than that of previous works. Our implemented CMC~\cite{tian2019contrastive} used an optical flow stream and two models for different views, obtaining 26.4\% points at top-1 accuracy. By using our proposed method with residual frames as the second view, the performance was improved to 36.5\%, which is 10.8\% points higher than the current state-of-the-art results. With optical flow data as the second view, this record could even reach 42.4\%. VCP~\cite{luo2020video} and VCOP~\cite{xu2019self} could reach a better performance than the numbers we picked in the table using R(2+1)D network, the best of which is 19.9\%. We did not include this here because we only use R3D as our network backbone. However, the performance of our worst model, which was 34.6\%, was still much better. In Fig.~\ref{retrieval}, qualitative results also show superiority of IIC compared with the baseline.

The transferability of the trained model was also tested on the HMDB51 dataset. The results are shown in Table~\ref{fig:retrieval hmdb}. A similar conclusion can be drawn. If only RGB video clips were used in the contrastive learning framework, the top-1 accuracy was 10.8\%. This performance was even 0.6\% higher than our implementation of the CMC method. This may have been caused by the effectiveness of 3D ConvNets. Without optical flow data, the performance of our proposed method is 13.4\% and 13.2\% respectively for the two different intra-negative sample generation strategies, revealing that inter-sample learning with intra-negatives is a good approach for self-supervised spatio-temporal feature learning, and is also good for unseen datasets. The best performance was obtained when optical flow data were treated as the second view, where 19.7\% on top-1 accuracy was achieved, outperforming the current state-of-the-art results by VCP~\cite{luo2020video} over 12\% points. 

\begin{table}[t]
  \centering
  \caption{Video retrieval performance on HMDB51.}
  \scalebox{0.9}{
  \setlength{\tabcolsep}{0.8mm}{
  \begin{tabular}{lccccc}
  \hline
  Methods&top1(\%)&top5(\%)&top10(\%)&top20(\%)&top50(\%)  \\
      \hline
  R3D (random) &6.7&18.3&28.3&43.1&67.9 \\
  VCOP~\cite{xu2019self} & 7.6 & 22.9&34.4&48.8&68.9\\
  VCP~\cite{luo2020video} & 7.6& 24.4& 36.3 & 53.6& 76.4 \\
  \hline
  One-view & 10.8 & 26.2 & 40.1 & 54.3 & 74.9  \\ 
  CMC~\cite{tian2019contrastive} & 10.2 & 25.3 & 36.6 & 51.6 & 74.3\\
  \hline
  IIC (repeat + res) & \textbf{13.4} & 32.7 & 46.7 & 61.5 & \textbf{83.8} \\
  IIC (repeat + u) & 17.1 & 41.9 & 55.1 & 70.4 & 84.9 \\
  IIC (shuffle + res) & 13.2 & \textbf{32.9} & \textbf{47.3} & \textbf{62.8} & 81.9 \\
  IIC (shuffle + v) & \textbf{19.7} & \textbf{42.9} & \textbf{57.1} & \textbf{70.6} & \textbf{85.9} \\
  \hline
  \end{tabular}}
  }
  \label{fig:retrieval hmdb}
\end{table}

\subsection{Comparison: video recognition}
Video feature representation is usually learned by supervised learning for video recognition task. Here we used our proposed method as an initialization strategy and the models were fine-tuned on two benchmark datasets, UCF101~\cite{ucf101} and HMDB51~\cite{hmdb}. The comparisons are among self-supervised methods for fair comparison because supervised methods usually used a much larger dataset, Kinetics~\cite{kinetics}, together with labels to pre-train their models.

When training models using our proposed IIC, data from different views were set as inputs to the same network. The distributions of data from different views were different. We argue that even though the network could handle different kinds of input, it would become a bottleneck when fine-tuned with labeled data from different views. Therefore, we had two choices when fine-tuning our trained models on the action recognition datasets: 1. use RGB video clips; 2. use the same data modality as view 2.

Because we had several settings for self-supervised training, four option configurations were selected and these models were fine-tuned with data from the two different views separately. This small test was conducted on UCF101 \textit{split}~1 only, and the results are shown in Table~\ref{table:fine-tune mode}. 
As we can see from the table, fine-tuning with RGB inputs yielded a worse performance than using the modality of the second view. Then we compare our model with other self-supervised learning methods by fine-tuning our models with the modality of the second view.

\begin{table}[tb]
\centering
  \caption{Results for different fine-tuning modes.}
  \scalebox{0.9}{
  \setlength{\tabcolsep}{1mm}{
  \begin{tabular}{c c c c c c c}
      \hline
      \multirow{2}{*}{Type} & \multirow{2}{*}{view1} & \multirow{2}{*}{view2} & \multicolumn{2}{c}{fine-tuning mode 1} & \multicolumn{2}{c}{fine-tuning mode 2} \\
      \cline{4-7}
       & & & modality & accuracy & modality & accuracy \\
      \hline
      repeat & RGB & res & RGB & 61.6 & res & \textbf{71.8} \\
      repeat & RGB & u  & RGB & 59.8 & u & \textbf{73.5} \\
      shuffle & RGB & res & RGB & 61.2 & res & \textbf{74.9} \\
      shuffle & RGB & v  & RGB & 61.6 & v & \textbf{63.1} \\
      \hline
  \end{tabular}}}
  \label{table:fine-tune mode}
\end{table}

From Table~\ref{fig:state of the art}, we can see that the model with ImageNet pre-trained weights can had a better performance than most methods. Recent methods have achieved better performances than a random initialization strategy, which means temporal information has been embedded by self-supervised learning to some extent. By fine-tuning models with data from the second modality, IIC could outperform VCP~\cite{luo2020video} by at maximum 8.4\% points when using the same network architecture. Note that the best performance on UCF101 dataset, 74.4\% at the top-1 accuracy, was achieved without using optical flow data, and better than O3N~\cite{fernando2017self} which also used residual inputs. Though the best performance for video retrieval was achieved with optical flow data, it was not as good as residual frames when fine-tuning. This could have been caused by a bias when training during self-supervised learning. If we only used one model to handle all input data, RGB video clips, its intra-negatives and optical flow data, the model may have concatenated more on distinguishing inputs with similar distributions as RGB video clips, resulting in bad initialization when fine-tuning only using optical flow data. Retrieval with our joint strategy can eliminate this drawback. This phenomenon could be improved if we use different models to handle different modality of data, which is also an option in our proposed framework. We leave this as our further work. 

The transferability was again tested on the HMDB51 dataset, which is more complicated because this is not only transferable for different tasks, but also on different datasets. Table~\ref{fig:state of the art} shows that IIC could handle this situation. With residual frames, our model yielded 38.3\% on HMDB51, which is the best among those methods without optical flow data. The size of ImageNet is much larger than our pre-trained dataset, UCF101 \textit{split}~1, while better performance was achieved. This indicates that our proposed method can be set as a good initialization strategy for other video related tasks. With optical flow data, especially $u$ data, good results can also be obtained, outperforming VCP~\cite{luo2020video} by 5.3\% points.

\begin{table}
  \centering
  \caption{Comparison of action recognition accuracy on UCF101 and HMDB51 datasets. Results are averaged over three splits. $^*$ indicates results using the same network backbone, R3D. $^\dag$ indicates methods using optical flow.}
  \scalebox{1}{
  \setlength{\tabcolsep}{1.5mm}{
  \begin{tabular}{lcc}
  \hline
  Method&UCF101(\%)&HMDB51(\%)  \\
  \hline
  Jigsaw~\cite{noroozi2016unsupervised} & 51.5 & 22.5 \\
  O3N (res)~\cite{fernando2017self} & 60.3 & 32.5 \\
  OPN ~\cite{lee2017unsupervised} & 56.3 & 22.1\\
  B\"uchler~\cite{buchler2018improving} & 58.6 & 25.0 \\
  Mas~\cite{wang2019self} & 58.8 & 32.6\\
  Geometry~\cite{gan2018geometry} & 54.1 & 22.6 \\
  CrossLearn~\cite{sayed2018cross}$^\dag$ & 58.7 & 27.2 \\
  CMC (3 views)~\cite{tian2019contrastive}$^\dag$ & 59.1 & 26.7\\
  \hline
  R3D (random)~\cite{xu2019self}$^*$ &54.5&23.4\\
  ImageNet-inflated~\cite{kim2019self}$^*$ & 60.3 & 30.7\\
  3D ST-puzzle~\cite{kim2019self}$^*$ & 65.8 & 33.7\\
  VCOP~\cite{xu2019self}$^*$ &64.9& 29.5\\ 
  VCP~\cite{luo2020video}$^*$ & 66.0& 31.5\\ 
  \hline
  IIC (repeat + res)$^*$ & 72.8 & 35.3 \\
  IIC (repeat + u)$^*$$^\dag$ & 72.7 & 36.8 \\
  IIC (shuffle + res)$^*$ & \textbf{74.4} & \textbf{38.3} \\
  IIC (shuffle + v)$^*$$^\dag$ & 67.0 & 34.0 \\
  \hline
  \end{tabular}}}
  \label{fig:state of the art}
\end{table}

\section{Conclusions}
In this paper, we proposed IIC, a self-supervised method for video representation learning, to learn rich temporal features from videos. We utilized the advantages of intra- and inter-sample learning and trained a spatio-temporal convolution neural network (3D-CNN) with intra-negative samples in contrastive multiview coding. Two intra-negative sample generation functions were proposed which break the temporal relations in input video clips. Different view selection options were also experimented. The trained models had learnt video representation and were applied to two video tasks, video retrieval and video recognition. With only one model handling different inputs, we could apply a joint retrieval strategy and our results showed that our models could outperform other methods by a large margin on video retrieval task. Experiments on video recognition also indicated that our proposed method could help the model learn better video representation.

\begin{acks}
This work was partially financially supported by the Grants-in-Aid for Scientific Research Numbers JP19K20289 and JP18H03339 from JSPS.
\end{acks}

\bibliographystyle{ACM-Reference-Format}
\bibliography{sample-base}

\appendix
\section{Algorithm}
\label{alg}
\begin{algorithm}[ht]
  \caption{Training with inter-intra contrastive learning framework}
  model: $net$, 
  video view: $X^1 = \{x_1^1, ..., x_N^1\}$, $X^2 = \{x_1^2, ..., x_N^2\}$, \\
  video index: $i$, memory bank: $M^1, M^2, M^{neg}$\\
  \algcomment{
    \fontsize{7.2pt}{0em}\selectfont Line 3,6 are the main differences from the baseline method CMC~\cite{tian2019contrastive}.
  }
  \begin{algorithmic}[1]
    \For{each iteration} 
     \State $x_i^1, x_i^2, i = load(Dataloader)$
     \State Generate intra-negative samples $x_i^{neg} = f(x_i^1)$
     \State $v_i^1 = net(x_i^1)$, $v_i^2 = net(x_i^2)$, $v_i^{neg} = net(x_i^{neg})$
     \State Fetch weights $W^1, W^2, W^{neg} = fetch(M^1, M^2, M^{neg}, i)$
     \State Form non-parametric weights $W^1 = concat(W^1, W^{neg})$, $W^2 = concat(W^2, W^{neg})$
     \State $loss = contrastive\_loss(W^1, W^2, v_i^1, v_i^2)$
     \State Update $net$ with $loss$
     \State Update memory banks $M^1$, $M^2$, $M^{neg}$ with $v_i^1, v_i^2, v_i^{neg}$
    \EndFor
  \end{algorithmic}
\end{algorithm}

\end{document}